\documentclass[sigconf,authorversion,screen,nonacm]{acmart}  %
\usepackage{graphicx}
\usepackage[font=small]{caption}
\usepackage{subcaption}
\usepackage{enumitem}
\setlist[itemize]{leftmargin=*}
\setlist[enumerate]{leftmargin=*}
\usepackage{hyperref}

\usepackage{acmart-taps}
\usepackage{amsmath}
\usepackage{mathtools}
\usepackage{amsthm}
\usepackage{longtable}
\usepackage{booktabs,multirow,multicol}

\usepackage[capitalize, nameinlink, noabbrev]{cleveref}

\newcommand{\oursfull}{Corrective Language Annotations for Robust InFerence}
\newcommand{\ours}{\textsc{Clarify}}

\copyrightyear{2024}
\acmYear{2024}
\setcopyright{acmlicensed}
\acmConference[UIST '24]{The 37th Annual ACM Symposium on User Interface Software and Technology}{October 13--16, 2024}{Pittsburgh, PA, USA}
\acmBooktitle{The 37th Annual ACM Symposium on User Interface Software and Technology (UIST '24), October 13--16, 2024, Pittsburgh, PA, USA}
\acmDOI{10.1145/3654777.3676362}
\acmISBN{979-8-4007-0628-8/24/10}

\usepackage{fancyhdr}
\AtBeginDocument{%
  \addtolength{\footskip}{2.0\baselineskip}%
    \fancyfoot[L]{\textit{\textbf{Preprint.}}}%
}

\begin{document}

\title{Clarify: Improving Model Robustness With Natural Language Corrections}

\author{Yoonho Lee}
\email{yoonho@cs.stanford.edu}
\affiliation{%
  \institution{Stanford University}
  \country{USA}
}

\author{Michelle S. Lam}
\email{mlam4@cs.stanford.edu}
\affiliation{%
  \institution{Stanford University}
  \country{USA}
}

\author{Helena Vasconcelos}
\email{helenav@cs.stanford.edu}
\affiliation{%
  \institution{Stanford University}
  \country{USA}
}

\author{Michael S. Bernstein}
\email{msb@cs.stanford.edu}
\affiliation{%
  \institution{Stanford University}
  \country{USA}
}

\author{Chelsea Finn}
\email{cbfinn@stanford.edu}
\affiliation{%
  \institution{Stanford University}
  \country{USA}
}




\begin{abstract}

The standard way to teach models is by feeding them lots of data. 
However, this approach often teaches models incorrect ideas because they pick up on misleading signals in the data. 
To prevent such misconceptions, we must necessarily provide additional information beyond the training data. 
Prior methods incorporate additional instance-level supervision, such as labels for misleading features or additional labels for debiased data. 
However, such strategies require a large amount of labeler effort.
We hypothesize that people are good at providing textual feedback at the concept level, a capability that existing teaching frameworks do not leverage. 
We propose Clarify, a novel interface and method for interactively correcting model misconceptions. 
Through Clarify, users need only provide a short text description of a model's consistent failure patterns. 
Then, in an entirely automated way, we use such descriptions to improve the training process. 
Clarify is the first end-to-end system for user model correction.
Our user studies show that non-expert users can successfully describe model misconceptions via Clarify, leading to increased worst-case performance in two datasets. 
We additionally conduct a case study on a large-scale image dataset, ImageNet, using Clarify to find and rectify 31 novel hard subpopulations.
\end{abstract}

\maketitle

\section{Introduction}

\begin{figure*}
\centering
\includegraphics[width=\linewidth]{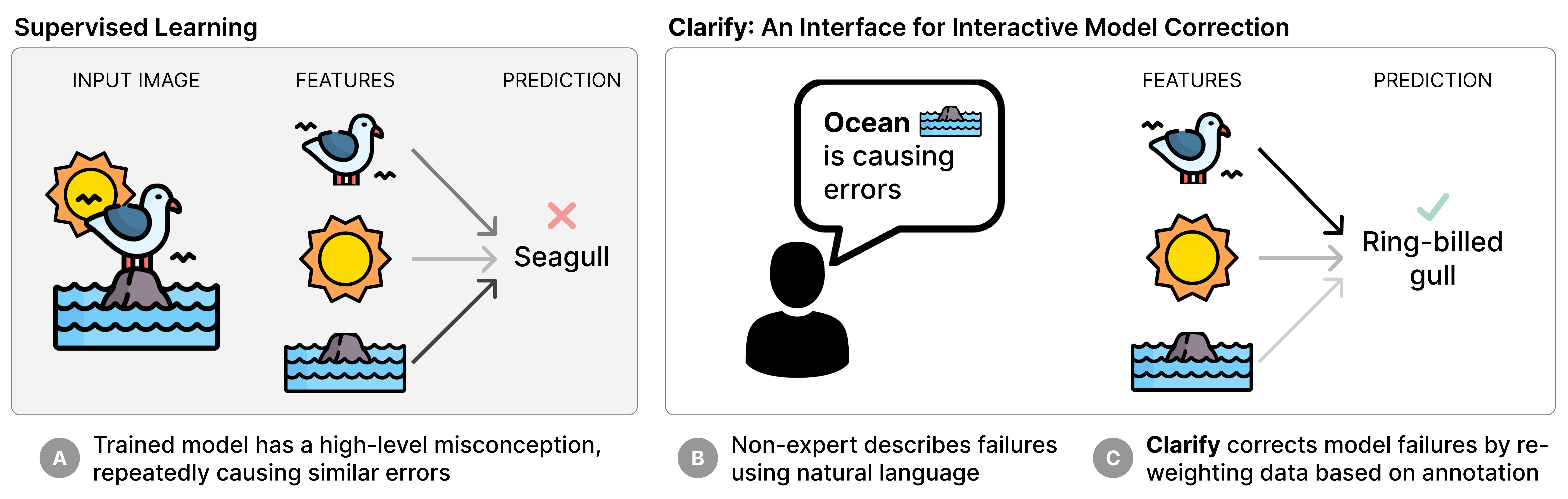}
\caption{
  \ours{} is an interface for interactively correcting model failures due to spurious correlations. (a) Given a model trained with supervised learning,
  (b) a human describes consistent failure modes of the model entirely in natural language.
  (c) We automatically incorporate these descriptions to improve the training process by reweighting the training data based on image-text similarity.
}
\label{fig:fig1}
\end{figure*}

Machine learning systems trained with supervised learning often learn high-level misconceptions.
For example, an image classifier trained to recognize birds may erroneously rely on background features like water rather than the visual appearance of the bird itself.
Such misconceptions can cause unexpected failures when the model is deployed in new environments, leading to poor performance on specific subpopulations~\citep{arpit2017closer,hendrycks2019benchmarking,shah2020pitfalls,geirhos2020shortcut,lazaridou2021mind,li_2022_whac_a_mole}. 
These misconceptions arise because models are trained to extract correlations from the training data, which may contain spurious or misleading signals.
Identifying such failure modes in advance is challenging due to the vast space of possible misconceptions.
Left unaddressed, such misconceptions can repeatedly cause the model to make similar errors, significantly degrading real-world performance.

Existing methods have sought to mitigate misconceptions by providing additional supervision beyond the training data.
Prior methods incorporate additional annotations about the spurious features, such as separate group indices indicating whether a bird image contains water, to encourage the model to ignore the spurious feature~\citep{sagawa2019distributionally}.
Alternatively, one can collect additional labeled data from a debiased distribution, for example, carefully curating images so that bird species is not correlated with the background~\citep{kirichenko2022last}.
A common theme in these approaches is that they require extensive human involvement in the form of additional instance-level supervision: in these approaches, additional annotations are needed at a scale comparable to that of the original training data.
This makes these strategies prohibitively costly for settings where the original training data is already close to the full annotation budget.
This is especially true in scenarios such as interactive machine learning~\cite{fails2003interactive,fogarty2008cueflik}, rapid model correction, or data-driven exploration.

We posit that far less supervision suffices if we provide targeted feedback at the level of \textit{concepts} rather than instances.
Targeted feedback is a cornerstone for robustness in various contexts outside teaching machine learning models.
Psychological studies underscore the pivotal role of corrective feedback in enhancing learning and decision-making~\citep{ilgen1979consequences,bangert1991instructional,kluger1996effects,hattie2007power}.
In causal inference, targeted interventions allow us to identify causal effects, going beyond the limitations of observational studies, which can only capture correlations~\citep{rubin1974estimating,pearl2009causality,scholkopf2021toward}.
Despite such insights, existing forms of annotation for robustness in supervised learning fall short in this regard: they lack the specificity of targeted feedback and are provided without knowledge of the actual behavior of naively trained models.
Through a lifetime of speaking and writing, people are highly adept at thinking and communicating at higher levels of abstraction.
However, existing frameworks for teaching models are not adequately designed to leverage people's ability to provide concept-level feedback.
This paper proposes a specific form of targeted feedback that aligns with these principles: natural language descriptions of model misconceptions.

We introduce~\oursfull{} (\ours{}), a novel system that allows users to interactively correct failures of image classifiers using natural language. 
We consider image classifiers obtained by fine-tuning pre-trained models such as CLIP~\citep{radford2021learning}. 
Although such classifiers achieve high average performance on held-out data, they often still suffer from high-level misconceptions.
\ours{} consists of an interface for collecting human feedback and a method for automatically incorporating this feedback to improve the training process.
During interactions with the system, users observe a trained model's predictions on a held-out dataset and write short text descriptions that identify consistent failure modes.
For instance, for a bird classifier relying on a spurious correlation between bird species and their backgrounds, a user might succinctly write that the model is mistakenly focusing on the ``water background''.
We note that our system diverges substantially from standard supervised learning: we collect annotations \textit{after} initial training and use these annotations in an entirely automated way to re-train the model based on the feedback.
Please refer to \cref{fig:fig1} for an overview of \ours{} in relation to traditional supervised learning, and \cref{fig:fig2} for a visualization of key interface features.

We instantiate \ours{} in a web app implementation to carry out non-expert user studies (N=26) and evaluate the gathered feedback in addition to re-trained models.
We find that within just a few minutes of interaction, non-expert users could use \ours{} to identify consistent failure modes of models trained with standard supervised learning.
Incorporating this feedback into the training process yields a statistically significant improvement in robustness: an average $17.1\%$ increase in the accuracy of the worst-performing subpopulations.
To further explore the ceiling of performance gains with \ours{}, we perform a case study on a large and diverse dataset, ImageNet, using an expert annotator.
This case study goes beyond standard datasets for spurious correlations with known failure modes and entails discovering and correcting previously unknown issues in a public dataset.
We were able to identify 31 novel hard subpopulations in the dataset. 
We leveraged this information to improve the average worst-case accuracy across these subpopulations from $21.1\%$ to $28.7\%$ with only a $0.2\%$ drop in average accuracy.
With \ours{}, we demonstrate that non-expert users can train and correct models by directly talking with them---opening up new design space for more efficient and accessible ways to design machine learning systems.

\section{Related Work}
\label{sec:related_work}
Our work draws upon literature in machine learning and human-computer interaction on strategies to efficiently correct machine learning models---whether to reduce training and annotation effort, bolster model robustness, or combat harmful failures.

\subsection{ML Perspectives on Model Correction}
Model correction methods in the machine learning literature tend to focus on developing novel algorithms while leaving user-facing processes intact, primarily focusing on using available labeled data more effectively.

\textbf{Teaching ML models.}
As machine learning models require more and more resources to train, it becomes increasingly important to optimize the training process.
The \textit{machine teaching} literature aims to formalize the optimal training set for a given task and characterize its training complexity.
While well-studied~\citep{goldman1991complexity,druck2008learning,mintz2009distant,zhu2015machine,simard2017machine,zhu2018overview}, its application to large-scale models has been limited, likely due to the substantial annotation burden required to teach a model from scratch.

Supervised learning, the dominant paradigm for training task-specific models, requires explicit labels for each instance and shows diminishing returns from additional human effort.
Although active learning methods aim to reduce this annotation burden by selecting the most informative datapoints for labeling~\citep{lewis1995sequential,settles2009active}, they still require humans to label individual datapoints.
Our work proposes a new form of supervision that can rectify spurious correlations in labeled datasets: natural language descriptions of model errors.
This form of supervision operates at a higher level of abstraction, providing a more efficient way to teach models with minimal additional annotation effort.

\textbf{Robustness to spurious correlations.}
Models trained with standard supervised learning often exhibit a bias towards shortcut features---simple features that perform well on the training distribution yet fail to capture the underlying causal structure~\citep{arpit2017closer,shah2020pitfalls,geirhos2020shortcut,pezeshki2021gradient}.
Recent works have proposed methods to mitigate this issue, such as learning multiple functions consistent with the data~\citep{teney2022evading,pagliardini2022agree,lee2022diversify,taghanaki2022masktune} and reweighting instances to render shortcut features non-predictive~\citep{sagawa2019distributionally,nam2020learning,creager2021environment,kirichenko2022last}.
However, these approaches often entail significant overhead for additional supervision, such as group labels indicating spurious features or carefully curated data free of spurious correlations.
In contrast, \ours{} requires only a few natural language descriptions of model errors, which are substantially easier to collect, rendering it especially practical for addressing misconceptions in large datasets.

\textbf{Discovering failure modes.}
Our work builds upon a growing body of literature aimed at identifying and correcting model failure modes.
Previous works discover poorly-performing subsets of data~\citep{chen2021mandoline,bao2022learning,d2022spotlight}, devise methods to rectify specific failures~\citep{santurkar2021editing,mitchell2021fast,yao2021refining,jain2022distilling}, or perform counterfactual data augmentation to penalize model reliance on erroneous features~\citep{kaushik2019learning,wu2021polyjuice,ross2021tailor,veitch2021counterfactual,vendrow2023dataset}.
More closely related to our work are methods that leverage vision-language models to describe failure modes with natural language~\citep{eyuboglu2022domino,wiles2022discovering,dunlap2022using,zhang2023diagnosing,Neuhaus_2023_ICCV,kim2023explaining}.
Natural language descriptions of error slices have the advantage of being interpretable and naturally grounded in human understanding.
However, many of the descriptions generated by these fully automated methods do not correspond to true model failures. 
For example, \citet{zhang2023diagnosing} reports that DOMINO~\citep{eyuboglu2022domino} can make nonsensical descriptions such as ``mammoth'' for a bird classification task.
Our approach avoids such errors by incorporating humans in the loop, making it possible to discover spurious correlations in large datasets such as ImageNet.

\subsection{Interactive Approaches to Model Correction}
Meanwhile, the HCI literature tends to approach model correction by leaving existing algorithms largely intact, but amplifying user involvement through new interactions and visualizations.

\textbf{Making ML models more accessible.}
The concept of a "low threshold" motivates HCI research on building systems accessible to non-experts~\citep{myers2000past}.
Many works have specifically focused on lowering the bar for end-users' participation in various stages of creating and using machine learning models.
Prior works have built tools for end-user data exploration~\citep{pair2017facets,tenney2020language,lam2024concept},
labeling~\citep{ratner2017snorkel,gao2022adaptive},
feature selection~\citep{fails2003interactive,dudley2018review},
model training~\citep{fogarty2008cueflik,carney2020teachable,matas2020lobe,ng2020understanding,ramos2020interactive,lane2021machine,azuremlstudio}, 
prompt engineering~\citep{swanson2021story,jiang2022prompt,tongshuang2022promptchainer},
and model auditing~\citep{suh2019anchorviz,wexler2020what,shen2021everyday,cabrera2021discovering,devos2022toward,lam2022end,deng2023understanding,yang2023beyond,cabrera2023zeno}.
In line with this rich literature, our work aims to enable non-expert end users to correct high-level misconceptions in machine learning models.
Since there is often a high effort barrier for users to engage in model development, we sought to demonstrate the efficacy of our approach even for limited amounts of user input.
To our best knowledge, \ours{} is the first to enable non-experts to use natural language to improve models in a fully end-to-end manner.

\textbf{Interactive ML.}
The field of interactive machine learning (IML) demonstrated that by engaging users in the model development process through interactive labeling, users could rapidly develop models that better aligned with their needs~\cite{fails2003interactive,amershi_powerToThePeople, francoise2021marcelle, dudley2018review}. Subsequent work on interactive machine teaching (IMT) further explored how users could act as teachers rather than just low-level data labelers~\cite{ramos2020IMT, simard2017IMT}. Both of these literatures have explored how to instantiate high-level \textit{concepts} with user-selected examples and demonstrations as well as predefined features and knowledge bases~\cite{fogarty2008cueflik, wekinator, ramos2020IMT, brooks2015featureInsight}. Building on this body of work, we find that concepts are a helpful level of abstraction for non-technical users, as they align well with how users tend to decompose and communicate knowledge~\cite{ng_IMTknowledgeDecomp}. \ours{} goes further by allowing users to specify arbitrary concepts in natural language to repair existing image classifiers that \textit{do not} already use concept-based abstractions.

\textbf{Eliciting high-level concepts.}
In our view, the most closely related works are those that elicit high-level concepts from humans~\citep{stretcu2023agile,lam2023model}.
However, a key difference between these works and ours is that we focus on \textit{negative knowledge}---teaching the model what not to learn---as opposed to these works, which specify what features the model should use.
Especially for intuitive tasks like image classification, user knowledge is often \textit{tacit} rather than explicit, making it hard to specify precisely~\citep{polanyi2009tacit}. 
Thus, it is easier for annotators to describe the failures of an existing model rather than define the desired behavior upfront.
Restricting the feedback to negative knowledge is also important for scalability, as it is much easier to identify a few failure modes in an otherwise well-performing model than to specify the full set of useful concepts.
This scalability is crucial for correcting spurious correlations in large-scale datasets such as ImageNet.

\section{\ours{}: A Natural Language Interface for Model Correction}
\begin{figure*}
  \centering
  \includegraphics[width=\linewidth]{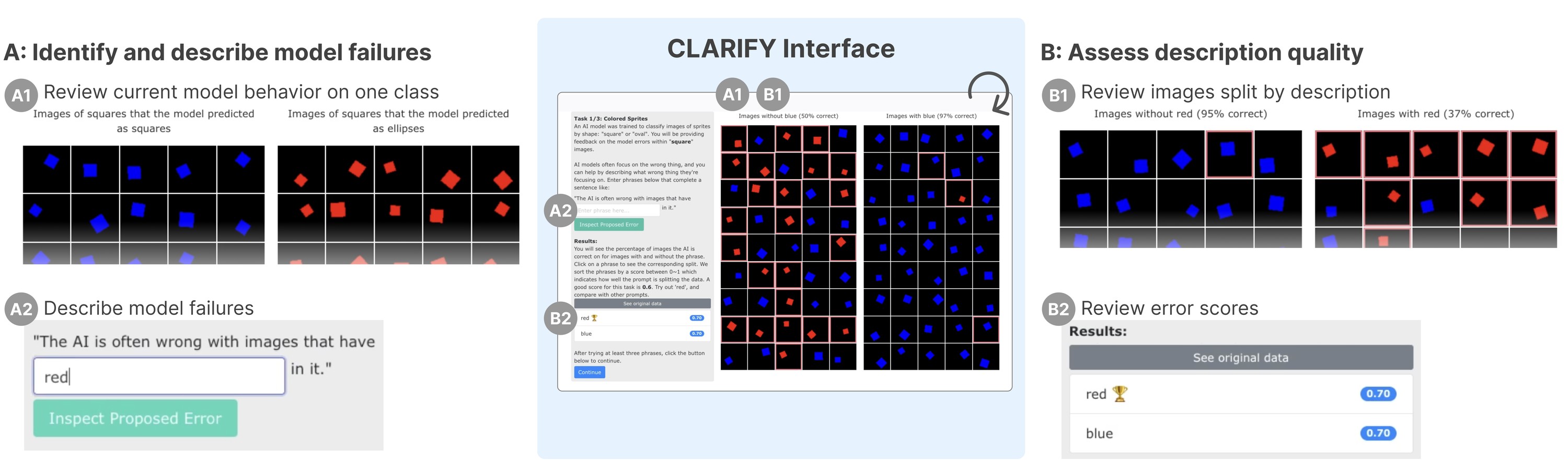}
  \caption{
    The \ours{} interface enables users to iteratively (A)~identify and describe model failures and (B)~assess the quality of these descriptions. Users can review image examples of correct and incorrect predictions on one class, such as ``square'' (A1). Based on observed differences, they can input short, natural language descriptions of model failures, such as ``red'' squares (A2). The system surfaces feedback by splitting the data using the provided description (B1) and displaying an error score (B2). Users can repeat the process to generate improved descriptions.
  }
  \label{fig:fig2}
\end{figure*}

\label{sec:method}

We now describe \oursfull{} (\ours), a novel system for identifying and mitigating spurious correlations in models trained with supervised learning.
The main idea behind \ours{} is to allow users to provide targeted natural language feedback to a model, helping the model focus on relevant features and ignore spurious ones.
We employ a natural language interface to facilitate this process, which we describe in detail in this section.
First, we describe the problem setting in~\cref{subsec:problem_setup}. 
We then describe a concrete example of an interaction with the interface in \cref{subsec:interface}, and two methods for incorporating this feedback into the training process in \cref{subsec:retraining}.

\subsection{Supervised Learning Problem Setup}
\label{subsec:problem_setup}
We consider a standard supervised learning setting, where we are given a dataset $\mathcal{D} = \{ (x_i, y_i) \}_{i=1}^{N}$ of $N$ labeled samples.
Each label $y_i$ belongs to one of $C$ different classes: $y_i \in \{1, \ldots, C\}$.
A model is trained to minimize the average loss across the training set, i.e., $\frac{1}{N} \sum_{i=1}^{N} \ell (f(x_i; \theta), y_i)$, where $\ell$ is a pointwise loss function such as cross-entropy, $f$ is the model, and $\theta$ denotes model parameters.
However, the dataset may inadvertently contain spurious correlations that hinder the model's ability to generalize to new distributions.
To formalize spurious correlations, we can consider an extended dataset that includes an unknown attribute $s_i$ for each instance, resulting in $\{ (x_i, y_i, s_i) \}_{i=1}^{N}$ where $s_i \in \{1, \ldots, S\}$.
For example, for a task where the labels $y_i$ are bird species, the spurious attributes $s_i$ could correspond to the background of the image $x_i$, which would be easier to infer from the input than the true label (i.e., bird species).
A model trained on $\mathcal{D}$ may learn to rely on $s_i$ to make predictions, thereby failing on new distributions where the previous correlation between $s_i$ and $y_i$ no longer holds.
In general, we do not have annotations for these spurious attributes $s_i$ or even know what they are in advance.
Our goal is to correct the model's reliance on these spurious attributes without knowing a priori what they are.

\subsection{Measuring Image-Text Similarity}
To describe spurious attributes given only class-labeled image data, we leverage the capabilities of multimodal models such as CLIP~\citep{radford2021learning}, which encodes images and text into a shared embedding space.
For a given image input $I$ and text input $T$, CLIP outputs representations from separate vision and language branches, $e_i = f_i(I)$ and $e_t = f_t(T)$, respectively.
This model is trained to maximize the similarity between the image and text representations for corresponding image-text pairs and minimize it for non-corresponding pairs through a contrastive loss function.
We can estimate the similarity between a pair of image and text inputs by computing the cosine similarity of their respective representations:
\begin{equation}
\textrm{sim}(I, T) = \frac{e_i \cdot e_t}{\|e_i\| \|e_t\|}.
\end{equation}
This black-box similarity function allows us to determine the relevance of a given image and text pair.
The next section describes how \ours{} leverages this relevance function to mitigate spurious correlations based solely on natural language feedback on a labeled validation set.

\begin{table*}[!t]
\label{table:nonExpert-phrases}
\centering
\footnotesize
\begin{tabular}{p{0.1\linewidth}  p{0.4\linewidth}  p{0.4\linewidth}}
\toprule
\textbf{Phrase Category} & \textbf{Waterbirds} & \textbf{CelebA} \\
\midrule
Best WGA (per-user) & a bird with no head or as landbirds and a red outline, a blurry vision and they don't look like real birds, artic birds, beak, bird swims water, dark backgrounds and tall trees, forest, forest, forest, forest, forest, forests, forests, grass, greenery, landscape, landscapes, leaves, no water, plants, red, sandy beaches, seagulls, seagulls, trees, water, water & any other hair color than blonde or light hair color, backgrounds, bleach blonds, brown hair, buns, curls, curly hair, dyed hair, females, glasses, light background, light colors, men, men, men or short hair, older women, orange hair, pink, red, red, short hair, short hair, short haired men, smiles, white backgrounds, women \\[0.1cm]
\hline
Best Error Score (per-user) & artic birds, birds, dark backgrounds and tall trees, ducks, forest, forest, forest, forest, forest, forests, forests, greenery, landscape, motion blur or can't make out a real bird, plants, sandy beaches, seagulls, trees, trees, trees, trees, water, water, water, waterfowl, wings & any other hair color than blonde or light hair color, bleach blonds, blond highlights, brown, darker blond hair, darker blonde, darker blonde hair, darker than blond, females, grey, males, men, men, men, men, men, men, men, men, men, men, men, pink, short white hair, very short hair, white, white, white \\[0.1cm]
\hline
All Others & a lot of dark colors and no blue water, a lot of tree trunks, aqua blue water, been generated by ai, bird, bird wading in water, birds, birds floating, birds floating in water, birds standing in water, birds water, black, blue, blue, branches, branches, dark backgrounds, dark backgrounds and small birds, dark colors, darker backgrounds and a lot of trees, extended wings, eyes, flightless birds, flowers, game birds, grass, green, green, green, green, green plants, humans, land, landscapes, length of leg, lots of tree trunks, more dark colors than light colors and a lot of trees, mountains, no water, no water, no water and dark backgrounds, ocean coasts, people, people, people, plants, reeds, seagulls, shadows, sticks, tree trunk, trees, trees, trees, trees, trees, trees, very dark backgrounds and a lot of trees, water plants, wings, woods & bad lighting, bangs, beards, black hair, blue, blue, blue background, blue or black, brown, brown or dark hair, dark hair, darker hair, dim lighting, fair hair, flaxen, gold, golden hair, hair, hair, hats, hats, hats or bows, hazy, letters, light hair, little visible hair, long hair, males, males, males, men, more dark colors than light colors, non-blond hair. dark hair color. not blond, nondarkened hair, not blond, orange hair, people not facing the camera, red hair, red hair, redheads, short, short hair, short or curly hair, short or pulled back hair, shoulders, signs, skin color that is similar to their hair color, smiles, smiling faces, sunglasses, tan skin, teenagers, teeth, very tan skin, women \\
\bottomrule
\end{tabular}
\caption{The full set of model failure description phrases provided by non-expert annotators in our user study. The ``Best WGA'' and ``Best Error Score'' phrases were selected by identifying the phrase that achieved the highest Worst-Group Accuracy or Error Score, respectively, for each participant.}
\end{table*}

\begin{figure*}
\centering \includegraphics[width=0.95\linewidth]{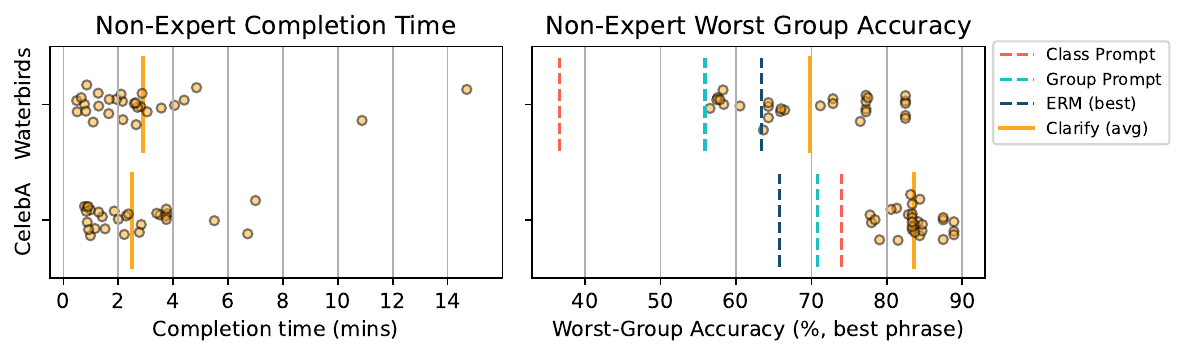}
\vspace{-3mm}
\caption{For both datasets, (left) non-experts completed annotation tasks using \ours{} in less than 3 minutes on average,
and (right) models retrained with non-expert annotations outperformed existing baselines in worst-group accuracy.}
\label{fig:nonexpert-time-wga}
\vspace{-3mm}
\end{figure*}
\begin{figure}
\centering \includegraphics[width=0.8\linewidth]{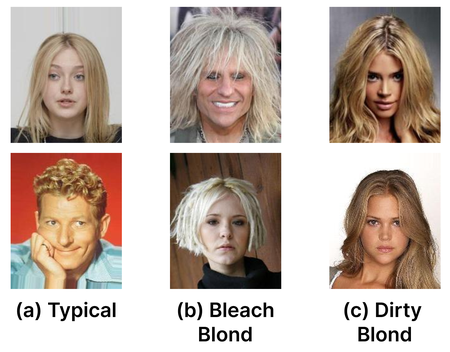}
\vspace{-3mm}
\caption{%
(a) Typical images from the ``blond'' class of CelebA.
Non-experts provided textual feedback corresponding to hard subpopulations of (b) lighter and (c) darker hair colors.
}
\label{fig:blond-subpopulations}
\end{figure}

\subsection{Interaction Workflow}
\label{subsec:interface}

To demonstrate how \ours{} enables non-expert users to correct model misconceptions, we will walk through a user's workflow with the system (\cref{fig:fig2}).
We will use a running example of a model trained to classify images of sprites as squares or ovals but mistakenly focuses on color rather than shape.

\textbf{Reviewing model behavior}. 
First, the user is presented with a summary view of the model's current behavior. 
The goal of this interface is to scaffold the user in rapidly identifying reasons underlying model failures.
Drawing from a validation dataset, we display one class at a time (i.e., images of squares) and divide the examples into those that the model correctly classified (i.e., images classified as squares) on the left versus those that it incorrectly classified (i.e., images classified as ovals) on the right (\cref{fig:fig2},~A1). 
By presenting the images in this way, \ours{} streamlines the user's task to one of identifying differences between sets. 
In our example, all of the images on the page are indeed squares, but the model is only making accurate predictions for the examples on the left and not those on the right. 
Comparing the images on the two sides, the user notices that the correct cases contain blue squares while the incorrect cases contain red squares.

\textbf{Describing model failures}.
Now that the user has an initial idea of the model's misconception, they are tasked with describing this failure mode. Our system accepts short natural language descriptions of model failures (\cref{fig:fig2},~A2). 
In particular, users are asked to complete the following fill-in-the-blank sentence: ``The AI is often wrong on images that have \_\_\_ in it.''
We find that this question is effective since users may not be familiar with the general concept of model failures or features.
Continuing our running example, the user enters the phrase ``red'' here to describe what they observed.

\textbf{Assessing descriptions}.
After the user submits their failure mode description, the \ours{} interface helps them assess whether the description effectively describes the model's misconception.
The system uses the CLIP model to compute the image-text similarity between each validation image and the user's description.
Images with a similarity score above a threshold are considered to contain the feature described by the user.
The interface presents a summary visualization that partitions the validation dataset based on this threshold, with matching images on the right and non-matching images on the left (\cref{fig:fig2},~B1).
Additionally, we display a $0-1$ score that indicates how well the description separates the error cases from the correct predictions (\cref{fig:fig2},~B2).
We note that while the interface only shows \textit{validation data} using the provided description, the user's natural language annotation will later be incorporated to partition the \textit{training data} for model retraining.

\textbf{Iterating on descriptions}.
As users may not be successful on their first attempt, \ours{} aids users in iterating on their descriptions. 
Descriptions can fail for two reasons: (1)~the description may be a valid differentiator, but may be modeled inaccurately due to the user's word choice and the limitations of CLIP-based similarity scoring, or (2)~the description may not sufficiently differentiate the correct and incorrect cases.
\ours{} allows users to identify both of these failure modes. 
For example, the user can see if the model is not accurately identifying images with the color red based on the keyword ``red'' alone.
In this case, they can experiment with alternate keywords to better isolate the difference, such as ``red square'' or ``crimson''.
After iterating and isolating the red examples, the user can see if the provided score is still low, indicating that this description is not sufficient to repair model errors.
With this information, users can revisit the original view and brainstorm additional descriptions, such as phrases related to the size or orientation of sprites.

We describe other details about the interface in~\cref{sec:app_interface}, including additional features that help users to refine their descriptions and assess their effectiveness.
In~\cref{subsec:exp_nonexpert}, we evaluate the performance of non-expert annotators using \ours{} and demonstrate that they can identify and describe model misconceptions.

\subsection{Automatic Fine-Tuning}
\label{subsec:retraining}
After collecting textual feedback from users, we incorporate this feedback into the training process for fine-tuning a foundation model.
While the strategy below applies to any form of training, in this paper, we consider fine-tuning only the last layer on top of a pre-trained backbone network with frozen parameters.
An error annotation is a tuple $(c, T, \tau)$, where $c$ is the class label, $T$ is the textual description, and $\tau$ is a threshold on the similarity function.
Given such an error annotation, we partition the training data within class $c$ into two subsets: $D_{>} = \{ (x_i, y_i) \mid \textrm{sim}(x_i, T) > \tau \}$ and $D_{<} = \{ (x_i, y_i) \mid \textrm{sim}(x_i, T) \leq \tau \}$.
These two subsets correspond to images that are more and less similar to the provided text prompt, respectively, and serve as indicators of the model misconception identified by the annotator.
Having identified these two subsets, we want to train a final model insensitive to the identified misconception, i.e., to achieve low training loss \textit{without} using the feature that separates the two subsets.

We propose to use a simple distributionally robust optimization (DRO) objective function to achieve this goal. 
Having identified the two subsets $D_{>}$ and $D_{<}$, we propose to minimize the maximum loss over the two subsets to achieve robustness to the identified misconception; the loss function is given by:
\begin{align}
  \max \left( \mathcal{L}(f_\theta, D_{>}), \mathcal{L}(f_\theta, D_{<}) \right),
\end{align}
where $\mathcal{L}(f_\theta, D)$ is the average loss over the subset $D$.
This objective ensures the model performs well on both subsets, avoiding the previous reliance on the spurious attribute.
We optimize this objective using stochastic gradient descent with the $\max$ operator computed for each minibatch.
We use this objective to train the last layer on top of a frozen pre-trained backbone model.
In~\cref{sec:experiments}, we will measure the effectiveness of this fine-tuning approach based on language feedback.
We note that this stage is fully automated, and there are no additional hyperparameters to tune beyond what was in the original training process.

\section{Evaluation}
\label{sec:experiments}

First, we note that our setup diverges substantially from assumptions in traditional supervised learning.
\ours{} involves collecting annotations \textit{after} an initial round of training, and these annotations consist of targeted concept-level feedback rather than model-agnostic instance-level feedback.
We consider this deviation from the conventional setup necessary to efficiently address the challenge of learning robust prediction rules from observational data.
We seek to empirically answer the following questions about the \ours{} system for interactively correcting model errors:
\begin{enumerate}
  \item Can non-expert users use \ours{} to identify and describe spurious correlations in models trained with supervised learning?
  \item Can \ours{} discover and rectify novel spurious correlations in large datasets such as ImageNet?
  \item How does \ours{} compare to various automated methods which do not require human feedback?
\end{enumerate}
For detailed experimental setup including datasets, models, and Prolific participants, see~\cref{sec:experiment_setup}.

\subsection{User Study: Non-Expert Annotators Can Describe Model Errors}
\label{subsec:exp_nonexpert}

Identifying and annotating spurious correlations is a more nuanced task than conventional forms of annotation such as class labeling.
This raises the question of whether non-expert annotators can perform this task, and if so, how efficiently they can do so.
To answer these questions, we conduct a user study (N=26) to assess the ability of non-expert users to identify and describe spurious correlations in models trained with supervised learning (see \cref{sec:experiment_setup} for study details).

In this study, we asked each participant to interact with models trained on the Waterbirds and CelebA datasets using the \ours{} interface.
In addition to qualitatively examining the feedback provided by participants, we evaluate the performance of the models trained on the feedback.
We measure the robustness of these re-trained models using the worst-group accuracy (WGA) metric, which measures the accuracy of the worst-performing subpopulation in the dataset.
For example, in the Waterbirds dataset, the worst-group accuracy is the minimum average accuracy across the four subpopulations (``landbird in land'', ``landbird in water'', ``waterbird in land'', and ``waterbird in water'').

We summarize the results of our user study in~\cref{fig:nonexpert-time-wga} and \cref{table:main_results}, comparing to zero-shot prompting and fine-tuning baselines.
Users were able to achieve these performance improvements with minimal annotation effort, averaging $2.7$ minutes (SD=$2.5$) per dataset.
As a point of comparison, \citet{chang2022an} found that annotators required $4.4$ and $10.2$ minutes to provide high-quality labels for $100$ examples in the much simpler MNIST and K-MNIST datasets, respectively.
For the best-performing annotation from each user, the average worst-group accuracy was $69.8$ (SD=$9.0$, max=$82.5$) for Waterbirds and $83.6$ (SD=$3.1$, max=$88.9$) for the CelebA dataset.
Single-factor ANOVAs showed a statistically significant increase in worst-group accuracy from using \ours{}: $\textrm{F}(1, 24) = 12.96, p<0.002$ for Waterbirds and $\textrm{F}(1, 24) = 262.44, p<0.001$.
We then conducted Tukey HSD posthoc tests to compare conditions between all pairs of methods. 
This test confirmed statistical significance at the $p<.005$ level for all pairwise between \ours{} and the baselines for both datasets.
In summary, non-expert annotations using \ours{} significantly outperformed the baseline methods we considered.

We additionally find that non-expert annotators identified previously unknown spurious correlations in the CelebA dataset.
In addition to the known spurious correlation between hair color and gender, participants identified subpopulations of ``dirty blonde'' and ``bleach blond'' individuals, which models consistently misclassified (\cref{fig:blond-subpopulations}).
Our findings suggest that \ours{} can enable non-expert annotators to identify and describe model misconceptions.
This opens up the possibility of leveraging a broader workforce for interactively improving models trained on web-scale datasets such as ImageNet or LAION~\citep{deng2009imagenet,schuhmann2022laion}.

\begin{figure*}
\centering
\includegraphics[width=0.9\linewidth]{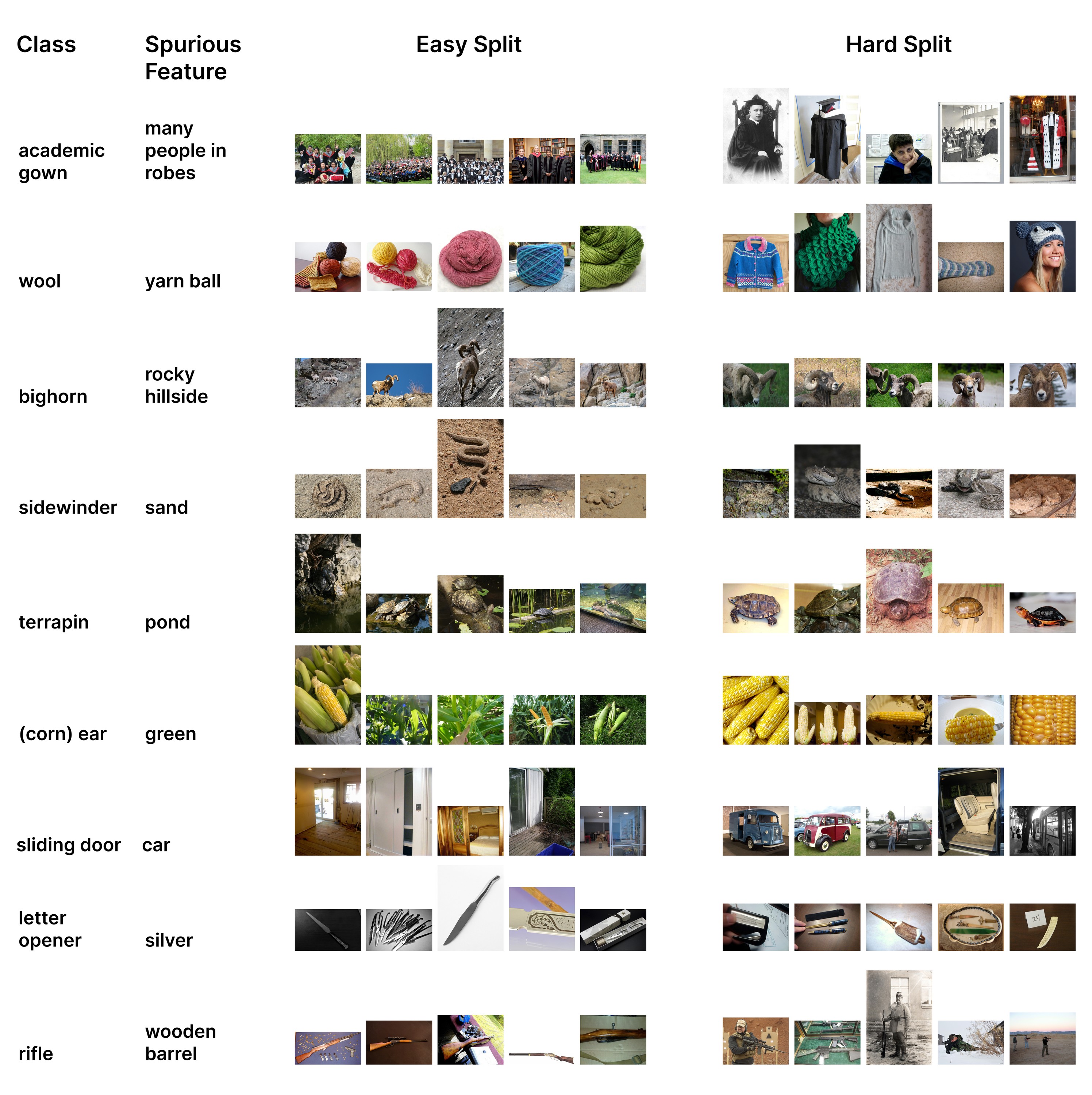}
\caption{
\label{fig:imagenet-all}
Representative samples corresponding to nine identified spurious correlations in ImageNet.
All images shown are in the ImageNet validation set, and belong to the class shown in the first column.
Similarity to the specified text annotation splits separates the ``easy'' and ``hard'' examples.
}
\end{figure*}

\aptLtoX{\begin{figure}
\includegraphics[width=0.95\linewidth]{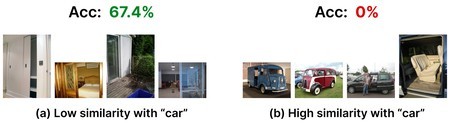}
\caption{\label{fig:imagenet-sliding-door}
An example of a spurious correlation found on ImageNet.
Within the ``sliding door'' class, the model successfully classifies (a) images of sliding doors inside buildings.
However, it is wrong on all instances of (b) sliding doors on cars.
This is one of the 31 spurious correlations we found; please refer to~\cref{fig:imagenet-all} for more visualizations.}
\end{figure}
\begin{figure}
\centering
\includegraphics[width=\linewidth]{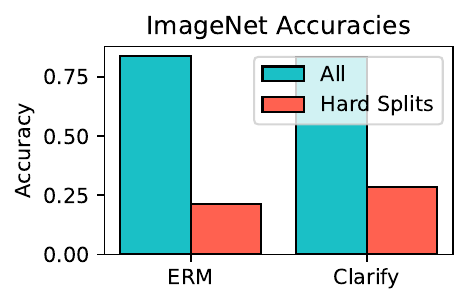}
\vspace{-8mm}
\caption{\label{fig:imagenet_accs}
Average accuracies on ImageNet data. Fine-tuning with Clarify substantially improves accuracy on hard splits, while keeping overall accuracy intact.}
\centering
\end{figure}}{\begin{figure*}
\begin{minipage}[b]{0.68\textwidth}
\centering
\includegraphics[width=0.95\linewidth]{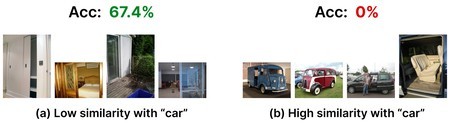}
\caption{\label{fig:imagenet-sliding-door}
An example of a spurious correlation found on ImageNet.
Within the ``sliding door'' class, the model successfully classifies (a) images of sliding doors inside buildings.
However, it is wrong on all instances of (b) sliding doors on cars.
This is one of the 31 spurious correlations we found; please refer to~\cref{fig:imagenet-all} for more visualizations.}
\end{minipage}
\hfill
\begin{minipage}[b]{0.29\textwidth}
\centering
\includegraphics[width=\linewidth]{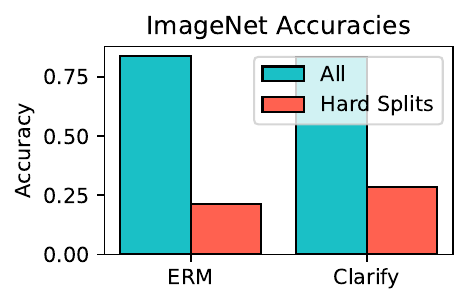}
\vspace{-8mm}
\caption{\label{fig:imagenet_accs}
Average accuracies on ImageNet data. Fine-tuning with Clarify substantially improves accuracy on hard splits, while keeping overall accuracy intact.}
\centering
\end{minipage}
\vspace{-2mm}
\end{figure*}}

\subsection{Case Study: Discovering and Mitigating Model Misconceptions in ImageNet}
\label{subsec:imagenet}

\begin{table}[h]
\centering
\label{tab:imagenet_annotations}
\begin{tabular}{|l|l|}
\hline
Class Name & Spurious Feature \\
\hline
cup & tea cup \\
weasel & snow weasel \\
wool & yarn ball \\
space bar & computer mouse \\
letter opener & silver \\
loupe & person holding a magnifying glass \\
mouse & desk and laptop \\
bakery & store front \\
sunscreen & person with sunburns \\
minivan & black minivan \\
plate rack & machine \\
briard & shaggy dog \\
lens cap & camera equipment \\
bighorn & rocky hillside \\
mushroom & red \\
rifle & wooden barrel \\
spotlight & shining \\
chocolate sauce & pastries with chocolate \\
terrapin & pond \\
sidewinder & sand \\
bikini & group of people \\
flatworm & coral reef \\
monitor & monitor on a desk \\
breastplate & museum display \\
projectile & rocket in a building \\
academic gown & many people in robes \\
velvet & pink velvet \\
bathtub & person \\
sliding door & car \\
partridge & tall grass \\
ear & green \\
\hline
\end{tabular}
\caption{The 31 identified spurious features in the ImageNet dataset.
All annotation was performed on the validation split.}
\end{table}

\begin{figure*}[t]
    \centering
    \includegraphics[width=\linewidth]{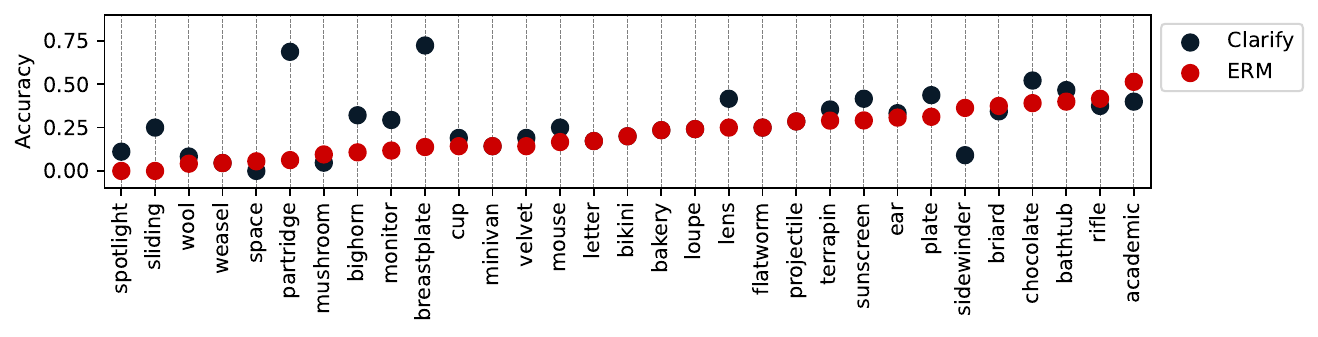}
    \caption{
    Average minority split accuracy for each of the 31 identified spurious correlations.
    Fine-tuning with textual feedback from \ours{} improves minority split accuracy in many classes.
    }
    \label{fig:imagenet-retrain}
\end{figure*}

We now evaluate whether \ours{} can be used to discover novel model misconceptions in models trained on the ImageNet training set.
It is important to develop tools to find consistent errors and methods to mitigate them for such widely used large-scale datasets.
For this evaluation, the authors of this paper use the \ours{} interface for ImageNet and additionally evaluate whether the resulting annotations can improve model robustness.

\textbf{Identified subpopulations.}
Using \ours{}, we identified 31 model misconceptions in ImageNet; we show a full list in~\cref{tab:imagenet_annotations} and visualize many of them in \cref*{fig:imagenet-all}
Despite ImageNet being a widely studied dataset, to our best knowledge, no prior works have identified these misconceptions.
For example, we visualize a spurious correlation in the ``sliding door'' class in~\cref{fig:imagenet-sliding-door}: sliding doors are negatively correlated with cars in the training set, causing standard models to misclassify cars with sliding doors.
We further evaluate the performance of a standard ImageNet model on each identified minority and majority split.
Results in~\cref{fig:imagenet-eval} show that a naively fine-tuned ERM model consistently underperforms on the minority split for each class, indicating that standard models indeed rely on each of these spurious correlations.
This trend continues to hold on ImageNet-V2, which follows a different distribution from the validation set we use to collect feedback.

\textbf{Fine-tuning while avoiding spurious correlations.} 
We use the collected annotations to fine-tune a model on ImageNet and evaluate this fine-tuned model on various splits of the ImageNet validation set.
Results in \cref{fig:imagenet-retrain} show that the retrained model achieves higher minority split performance on many classes.
Aggregate metrics in~\cref{fig:imagenet_accs} show that fine-tuning with \ours{} annotations improves the average minority-split accuracy from $21.1\%$ to $28.7\%$, with only a $0.2\%$ drop in overall average accuracy.
We emphasize that no additional data was used during fine-tuning---the annotations from \ours{} were only used to find a better reweighting of the same training data used to train the original ERM model.

\subsection{Technical Evaluation}
\label{subsec:automated}

We compare \ours{} with various automated methods for handling model misconceptions.
We note that none of the following comparisons are apples-to-apples: \ours{} operates in a new problem setting that involves targeted human feedback, while the automated methods are designed to use pre-existing data, annotations, or models.
Our goal is to see how the new form of targeted supervision in \ours{} compares to prior approaches since it is substantially more information-dense and expressive and thus scalable to large datasets.

\textbf{Comparison with zero-shot prompting methods.}
We compare the re-trained classifier using \ours{} annotations with zero-shot prompting methods in~\cref{tab:zero_shot}.
\ours{} shows substantially better worst-group accuracy and robustness gap on the Waterbirds and CelebA datasets.
Among these points of comparison, RoboShot~\citep{adila2023zero} is notable as it is an automated method that leverages state-of-the-art foundation models~\citep{jia2021scaling,chen2022altclip,OpenAI2023GPT4TR}.
We note that this is not necessarily a fair comparison in either direction: RoboShot uses a powerful language model to alter its prompts, whereas \ours{} leverages targeted human feedback.
Nevertheless, this comparison is still informative in that it shows that we can get much more leverage out of natural language feedback by having it directly address gaps in existing training data.

\begin{table*}[!t]
\centering
\begin{tabular}{llcccccc}
\toprule
&                      & \multicolumn{3}{c}{Waterbirds} & \multicolumn{3}{c}{CelebA} \\ 
\cmidrule(l){3-5} \cmidrule(l){6-8}
Data Assumptions & Method & WG & Avg & Gap & WG & Avg & Gap \\ 
\midrule
\multirow{2}{*}{Zero-Shot} 
  & Class Prompt            & 36.6 & 92.2 & 55.6 & 74.0 & 81.9 & 7.9 \\ 
& Group Prompt              & 55.9 & 87.8 & 31.9 & 70.8 & 82.6 & 11.8 \\ 
\cmidrule{2-8} 
\multirow{4}{*}{Labels}
  & ERM & 7.9  & 93.5 & 85.6 & 11.9 & 94.7 & 82.8 \\
& ERM (ours)               & 63.4 & 96.0 & 32.6 & 31.1 & 95.4 & 64.3 \\
& ERM (ours, class-balanced) & 48.6 & 95.2 & 46.7 & 65.8 & 93.4 & 27.6 \\
& ERM (ours, worst-class)  & 55.9 & 95.8 & 39.9 & 56.9 & 94.1 & 37.2 \\
\cmidrule{2-8}
\multirow{3}{*}{Labels, Text Feedback} 
& \ours{} (avg non-expert)     & 69.8 & 84.1 & 13.3 & 83.7 & 93.2 & 9.5 \\
& \ours{} (best non-expert)     & \textbf{82.5} & 90.7 & 8.2 & 88.8 & 92.9 & 4.1 \\
& \ours{} (author)         & 75.7 & 83.8 & \textbf{8.1} & \textbf{89.1} & 92.1 & \textbf{3.0} \\
\cmidrule{2-8} 
\multirow{4}{*}{Labels, Group Labels}
& DFR (downsample)        & 63.9 & 91.8 & 27.9 & 76.9 & 92.5 & 15.6 \\
& DFR (upsample)         & 51.3 & 92.4 & 41.1 & 89.6 & 91.8 & 2.2 \\
& DFR (our implementation)             & 78.7 & 90.8 & 12.1 & \textbf{90.6} & 91.9 & \textbf{1.3} \\
& Group DRO (our implementation)       & \textbf{81.3} & 88.1 & \textbf{6.8}  & 89.2 & 91.8 & 2.7 \\
\bottomrule
\end{tabular}
\caption{
\label{table:main_results}
Evaluation of methods for group robustness using the CLIP-ResNet50 backbone.
For \ours{}, we show the average and best participant from our non-expert user study (N=26) in addition to feedback from an author of this paper.
Fine-tuning with annotations from \ours{} consistently outperforms methods that use only text (zero-shot) or label information.
All results other than ours are from~\citet{zhang2022contrastive}.}
\end{table*}

\begin{table*}[t!]
\centering
\begin{tabular}{llcccccccccccc}
\toprule
\multirow{2}{*}{Model} & \multirow{2}{*}{Method} & \multicolumn{3}{c}{Waterbirds} & \multicolumn{3}{c}{CelebA} \\
\cmidrule(lr){3-5} \cmidrule(lr){6-8}
&& Avg & WG($\uparrow$) & Gap($\downarrow$) & AVG & WG($\uparrow$) & Gap($\downarrow$) \\
\toprule
\multirow{3}{*}{ALIGN}
& Class Prompt & 72.0 & \underline{50.3} & 21.7 & 81.8 & 77.2 & 4.6 \\
& Group Prompt & 72.5 & 5.8 & 66.7 & 78.3 & 67.4 & 10.9 \\
& RoboShot~\citep{adila2023zero} & 50.9 & 41.0 & \textbf{9.9} & 86.3 & \underline{83.4} & \underline{2.9} \\
\cmidrule{2-8}
\multirow{3}{*}{AltCLIP}
& Class Prompt & 90.1 & 35.8 & 54.3 & 82.3 & \underline{79.7} & \underline{2.6} \\
& Group Prompt & 82.4 & 29.4 & 53.0 & 82.3 & 79.0 & 3.3 \\
& RoboShot~\citep{adila2023zero} & 78.5 & \underline{54.8} & \underline{23.7} & 86.0 & 77.2 & 8.8 \\
\cmidrule{2-8}
\multirow{4}{*}{CLIP (ViT-L/14)}
& Class Prompt & 88.7 & 27.3 & 61.4 & 80.6 & 74.3 & 6.3 \\
& Group Prompt & 70.7 & 10.4 & 60.3 & 77.9 & 68.9 & 9.0 \\
& RoboShot~\citep{adila2023zero} & 79.9 & \underline{45.2} & 34.7 & 85.5 & \underline{82.6} & \underline{2.9} \\
\cmidrule{2-8}
& \ours{} & 96.8 & \textbf{81.8} & \underline{14.9} & 90.9 & \textbf{88.8} & \textbf{2.1} \\
\bottomrule
\end{tabular}
\caption{
\label{tab:zero_shot}
Comparison with different zero-shot CLIP prompting strategies for group robustness. 
Fine-tuning with \ours{} substantially outperforms RoboShot, a method that leverages state-of-the-art foundation models to automatically generate text prompts.
All results besides ours are from~\citet{adila2023zero}.
}
\end{table*}

\textbf{Comparison with methods for spurious correlations.}
We assess how re-training a model with expert annotations from \ours{} compares to existing automated methods for addressing spurious correlations.
We compare with representative prior methods, which similarly fine-tune CLIP models and/or reweight training data.
In addition to \ours{}, we evaluate zero-shot CLIP~\citep{radford2021learning} with class-based and group-based prompts, DFR~\citep{kirichenko2022last}, and Group DRO~\citep{sagawa2019distributionally}.
We describe experimental details for each method in~\cref{sec:experiment_setup}.
Our results on the Waterbirds and CelebA datasets, summarized in~\cref{table:main_results}, demonstrate that \ours{} consistently outperforms approaches that use zero-shot prompts or class labels in terms of worst-group accuracy and robustness gaps and is competitive with specialized methods that use instance-level annotations for spurious attributes.
We show extended results with an alternative network architecture as the pre-trained backbone in~\cref{table:main_results_full}.

Moreover, the key advantage of \ours{} is its scalability to large datasets, a feature that no prior automated method has demonstrated.
Such scalability is crucial when applying these ideas to real-world problems where the scale and diversity of data are ever-increasing.

\textbf{Comparison with automated bias discovery.}
Since annotation time is a key cost of \ours{}, we investigate whether such feedback brings concrete benefits over fully automated methods for discovering model biases.
Specifically, we compare \ours{} with Bias-to-Text~\citep{kim2023explaining}, a representative method for automated bias discovery.
We evaluated the automated Bias-to-Text pipeline on several classes in the ImageNet validation set, in which we identified spurious correlations. 
We find that Bias-to-Text can identify relevant keywords for each class, but it has specific pitfalls that make it difficult to use without human oversight.
In Table~\ref{tab:bias_to_text}, we show 10 keywords identified by Bias-to-Text for 6 classes in ImageNet. 
We note that the top identified keywords, i.e., the ones with the highest CLIP score, often describe something highly related to the class label, such as ``goat'' for the ``bighorn''.
Additionally, we numerically compare the discovered annotations in their ability to improve model robustness in Figure~\ref{fig:b2t_accs_comparison}.
The annotations for Bias-to-Text show substantially higher minority split accuracy (Clarify 21.1\%, Bias-to-Text 45.2\%), with a smaller gap with the majority split.
Furthermore, after re-training with these annotations, we observe a slight decrease in held-out minority split accuracy (45.2\% to 44.3\%).
This is in contrast to re-training with Clarify annotations, which substantially improved minority split accuracy (21.1\% to 28.7\%).
However, we note that automated discovery methods are highly useful in the context of \ours{}, as they can prime annotators with a set of candidate keywords or help prioritize the most promising parts of the dataset.

\section{Discussion}
\label{sec:discussion}

Our evaluation demonstrates that \ours{} enables non-expert users to identify and address misconceptions within machine learning models in an end-to-end manner.
User feedback is immediately actionable and can be used to improve model robustness without any additional data collection.
Allowing users to improve models end-to-end is important for both scalability and building public trust.
The worst-case accuracy gains from re-training with \ours{} annotations are substantial, and the method is competitive with or outperforms existing automated methods for addressing spurious correlations.
Furthermore, we show that \ours{} can be used to discover novel misconceptions from training on large-scale datasets such as ImageNet and that the resulting annotations can be used to improve model robustness.

\subsection{Comparison to Supervised Learning}
Our approach to teaching machines diverges substantially from standard supervised learning.
We highlight two key properties of \ours{} which offer complementary strengths to conventional supervised learning.
First, we collect annotations \textit{after} initial training, allowing the model's behavior to inform the annotation process.
Interacting with a model after training enables our interface to specifically elicit \textit{negative knowledge} from users, i.e., telling the model what \textit{not} to focus on.
People are often better at identifying errors than articulating complex rules, and negative knowledge can fill the gaps in the positive knowledge in the original labeled training set.
Second, annotations from \ours{} have a substantially higher information density than conventional forms of annotations.
Unlike instance-specific labels, textual feedback encapsulates concept-level ``global'' insights applicable across the entire dataset, making it a more efficient mode of human-AI communication.

\subsection{Limitations}
However, we acknowledge certain limitations of \ours{} in its current form.
While we find that non-expert users can provide meaningful feedback on model errors, eliciting high-quality feedback from non-experts remains challenging.
Their feedback can be too generic, e.g., ``the model is wrong'' or directly describing the class label rather than the spurious correlation; signaling the right level of granularity to the user is a challenge in this mode of interaction.
We also find that some users could only find model errors that are very visually salient, such as the presence of a specific large object in the image.
While many users provided useful feedback, some struggled, likely because they had no prior knowledge of the model's capabilities.

In general, \ours{} can only help with model failures that can be concisely described in natural language. 
This excludes more complex failures, such as those requiring domain-specific knowledge or not easily articulated in text.
Also, as much as \ours{} benefits from using a pre-trained backbone model for bridging between natural language descriptions and images, it inherits the limitations of the pre-trained CLIP model.
This includes biases in the training distribution and limited coverage of specialized domains such as medical imaging or scientific data.

\subsection{Future Work}
A natural extension of \ours{} is to apply this framework of collecting textual feedback on model errors to large language models (LLMs).
A key challenge in this direction is in designing effective user interfaces that allow users to quickly understand a model's overall behavior and develop strategies to elicit specific, actionable user feedback.
Enabling this mode of interaction, where users can teach LLMs by critiquing their mistakes, could be a promising approach for making powerful models easier to align and adapt to specific user and community needs without requiring extensive annotation costs.
Collecting feedback \textit{after} initial model training is especially appealing in the context of current LLMs, where the full scope of a model's emergent behaviors can be difficult to predict in advance.

We are also excited about increasing expressivity by designing interfaces that allow users to provide more nuanced and open-ended feedback, potentially through extended text passages or interactive multi-turn dialogue with the model.
Such interfaces could leverage more knowledge and expertise from people, including domain experts, which may be particularly valuable in specialized domains such as healthcare or scientific discovery.
We intentionally scoped our evaluation to non-experts to demonstrate the broad applicability of our end-to-end approach.
Our non-expert results establish a floor for improvement from the proposed workflow, and further iterations will only benefit from richer user feedback from users with more expertise.
Allowing for richer feedback will also likely be critical in more complex tasks like open-ended instruction following and in data modalities such as video or audio, where the model's behavior is more complex and harder to summarize succinctly.

\section{Conclusion}
\label{sec:conclusion}

This paper introduces \ours{}, a novel interface for correcting high-level model misconceptions in machine learning models.
\ours{} enables non-expert users to provide targeted feedback on model errors, which can be used to improve model robustness.
We also show that \ours{} can be used to discover novel misconceptions in large-scale datasets such as ImageNet.
We believe that the general idea of correcting models with targeted textual feedback has the potential to substantially improve model performance while reducing the need for extensive manual annotation.

\begin{acks}
We thank members of the IRIS lab and Pang Wei Koh for helpful discussions and feedback.
YL acknowledges support from KFAS.
This work was partly supported by NSF CAREER award 2237693 and Schmidt Sciences.
\end{acks}

\bibliographystyle{ACM-Reference-Format}
\bibliography{paper}

\appendix
\clearpage

\section{Additional Interface Details}
\label{sec:app_interface}
In this section, we provide additional details about the \ours{} interface, which we found helpful for eliciting natural language feedback from non-expert users.

\textbf{Error score.} 
The provided Error Score is a rough proxy for how well a given text description predicts model errors. 
We emphasize that this score is \textit{not} used in the training process and is only meant to give non-expert users a rough idea of what descriptions are useful.
It is computed as follows.
Consider input text prompt $T$, and let $D_{\textrm{correct}}$ and $D_{\textrm{error}}$ be subsets of the validation dataset for a given class that the model made correct and incorrect predictions on, respectively.
We denote the cosine similarities between the $T$ and the images in each subset as $S_{\textrm{correct}} = \{ \textrm{sim}(I, T) \mid I \in D_{\textrm{correct}} \}$ and $S_{\textrm{error}} = \{ \textrm{sim}(I, T) \mid I \in D_{\textrm{error}} \}$.
To quantify how well image similarity with $T$ can predict model errors, we compute the best class-balanced binary classification accuracy among similarity thresholds $\tau$.
Denoting this accuracy as $\textrm{Acc}_\tau$, the \textrm{error score} is computed as $2 \times (\textrm{Acc}_\tau - 0.5)$, so that uninformative prompts receive a score of 0 and prompts that perfectly predict model errors receive a score of 1.

\textbf{Similarity threshold.}
For each natural language threshold, we determine a similarity threshold $\tau$, which can be chosen by the user after inspecting the similarity scores for a representative sample of images or can be automatically chosen as the threshold that maximizes the Error Score.
For each class, only the textual feedback with the highest Error Score is used for retraining.
Together with this threshold, we can specify a spurious correlation using a tuple of the form (class label, text prompt, similarity threshold) corresponding to a binary classifier that predicts model errors on that class.

\textbf{Additional backend features for large datasets.}
We found that a few more optional features can help annotate spurious correlations in larger datasets like ImageNet. 
We begin by narrowing down the 1000 classes to a smaller number of classes (e.g., 100) most likely to have identifiable spurious correlations.
To do so, we first prune out classes with too low or too high accuracy (e.g. accuracy below 0.2 or above 0.8), to ensure a sufficient number of correct and incorrect predictions for each class.
For the remaining classes, we caption each image with an image captioning model \citep[BLIP]{li2022blip} and use a keyword extraction model \citep[KeyBERT]{grootendorst2020keybert} to suggest a pool of up to 50 keywords for each class, a procedure inspired by~\citet{kim2023explaining}.
Through \ours{}, we interact with the top 100 classes according to the maximum error score across the candidate keywords.
The user is shown the top 10 candidate keywords during interactions as a helpful starting point.
We expect that these features will similarly be helpful for other large datasets.

\section{Experimental Details}
\label{sec:experiment_setup}

\aptLtoX{\begin{table*}[]
\begin{tabular}{rll}
    \multicolumn{3}{c}{Spotlight class (ours: ``shining'')} \\
\hline
    Keyword &  CLIP Score & Subgroup Acc \\
\hline
street lamp &        3.32 &    0.0 (N=1) \\
       lamp &        2.50 &   66.7 (N=6) \\
        top &        2.46 &    0.0 (N=1) \\
    kitchen &        2.22 &   50.0 (N=2) \\
     street &        2.07 &   33.3 (N=3) \\
      suite &        1.94 &    0.0 (N=1) \\
       city &        1.87 &    0.0 (N=3) \\
       room &        1.77 &    0.0 (N=2) \\
      light &        1.51 &  81.8 (N=22) \\
      night &        1.12 &   80.0 (N=5) \\
\bottomrule
\end{tabular}
\begin{tabular}{rll}
        \multicolumn{3}{c}{Rifle class (ours: ``wooden barrel'')} \\
\hline
        Keyword &  CLIP Score & Subgroup Acc \\
\hline
         person &        1.59 &  14.3 (N=14) \\
        soldier &        0.60 &    0.0 (N=8) \\
project picture &        0.20 &    0.0 (N=3) \\
       soldiers &        0.18 &    0.0 (N=5) \\
      dark room &       -0.47 &    0.0 (N=1) \\
        machine &       -0.95 &   20.0 (N=5) \\
            gun &       -1.69 &  30.0 (N=20) \\
       nice gun &       -1.98 &    0.0 (N=2) \\
    machine gun &       -2.00 &   20.0 (N=5) \\
        weapons &       -2.01 &   42.9 (N=7) \\
\bottomrule
\end{tabular}
\begin{tabular}{rll}
\multicolumn{3}{c}{Academic Gown class (ours: ``many people in robes'')} \\
\hline
            Keyword &  CLIP Score & Subgroup Acc \\
\hline
             person &        0.51 &  25.0 (N=24) \\
              photo &        0.50 &  27.3 (N=11) \\
           graduate &        0.23 &  13.3 (N=15) \\
          graduates &        0.21 &   25.0 (N=8) \\
         graduation &        0.16 &  13.3 (N=15) \\
               pose &        0.12 &  20.0 (N=10) \\
              poses &        0.01 &    0.0 (N=4) \\
           students &       -0.25 &    0.0 (N=7) \\
graduation ceremony &       -0.57 &  16.7 (N=12) \\
           ceremony &       -1.27 &  23.1 (N=13) \\
\bottomrule
\end{tabular}
\begin{tabular}{rll}
        \multicolumn{3}{c}{Bighorn class (ours: ``rocky hillside'')} \\
\hline
             Keyword &  CLIP Score & Subgroup Acc \\
\hline
                goat &        1.39 &   5.9 (N=17) \\
               sheep &        1.35 &    0.0 (N=9) \\
       mountain goat &        0.11 &    0.0 (N=2) \\
          biological &        0.06 &   28.6 (N=7) \\
  biological species &        0.00 &   28.6 (N=7) \\
             species &       -0.03 &   28.6 (N=7) \\
       bighorn sheep &       -0.04 &    0.0 (N=4) \\
bighorn sheep stands &       -0.29 &    0.0 (N=2) \\
              stands &       -1.32 &    0.0 (N=7) \\
                herd &       -2.27 &   14.3 (N=7) \\
\bottomrule
\end{tabular}
\begin{tabular}{rll}
       \multicolumn{3}{c}{Loupe class (ours: ``person holding a magnifying glass'')} \\
\hline
         Keyword &  CLIP Score & Subgroup Acc \\
\hline
           black &        0.01 &    0.0 (N=4) \\
          camera &       -0.08 &   28.6 (N=7) \\
            book &       -0.65 &   33.3 (N=3) \\
         compact &       -0.72 &    0.0 (N=2) \\
  compact camera &       -1.12 &    0.0 (N=2) \\
           watch &       -1.50 &    0.0 (N=1) \\
          pocket &       -1.84 &    0.0 (N=1) \\
    pocket watch &       -2.05 &    0.0 (N=1) \\
           glass &       -2.30 &  57.1 (N=14) \\
magnifying glass &       -6.19 &   71.4 (N=7) \\
\bottomrule
\end{tabular}
\begin{tabular}{rll}
  \multicolumn{3}{c}{Weasel class (ours: ``snow weasel'')} \\
\hline
             Keyword &  CLIP Score & Subgroup Acc \\
\hline
       bear cub sits &        1.72 &    0.0 (N=1) \\
      black bear cub &        1.62 &   50.0 (N=2) \\
    young black bear &        0.81 &    0.0 (N=1) \\
  biological species &       -0.17 &  85.7 (N=14) \\
      dead squirrels &       -0.19 &    0.0 (N=1) \\
          file photo &       -0.19 &    0.0 (N=2) \\
        undated file &       -0.36 &    0.0 (N=2) \\
  undated file photo &       -0.41 &    0.0 (N=2) \\
               grass &       -0.87 &   85.7 (N=7) \\
squirrels were found &       -0.87 &    0.0 (N=1) \\
\bottomrule
\end{tabular}
  \caption{Comparison to Bias-to-Text~\citep{kim2023explaining}, an automated bias discovery method on ImageNet.
    We show the top 10 keywords identified by Bias-to-Text in descending order of their recommended score.
    We also show the text feedback provided through \ours{} for comparison.
    The keywords identified by Bias-to-Text often include irrelevant words or correspond to very small subpopulations, indicating that current automated methods ultimately require human oversight or intervention to discover the most relevant and biased subpopulations.}
  \label{tab:bias_to_text}
\end{table*}}{\begin{table*}[]
  \hfill \begin{minipage}{0.45\linewidth} \centering
\begin{tabular}{rll}
    \multicolumn{3}{c}{Spotlight class (ours: ``shining'')} \\
\toprule
    Keyword &  CLIP Score & Subgroup Acc \\
\midrule
street lamp &        3.32 &    0.0 (N=1) \\
       lamp &        2.50 &   66.7 (N=6) \\
        top &        2.46 &    0.0 (N=1) \\
    kitchen &        2.22 &   50.0 (N=2) \\
     street &        2.07 &   33.3 (N=3) \\
      suite &        1.94 &    0.0 (N=1) \\
       city &        1.87 &    0.0 (N=3) \\
       room &        1.77 &    0.0 (N=2) \\
      light &        1.51 &  81.8 (N=22) \\
      night &        1.12 &   80.0 (N=5) \\
\bottomrule
\end{tabular}
  \end{minipage} \hfill
  \begin{minipage}{0.45\linewidth} \centering
\begin{tabular}{rll}
        \multicolumn{3}{c}{Rifle class (ours: ``wooden barrel'')} \\
\toprule
        Keyword &  CLIP Score & Subgroup Acc \\
\midrule
         person &        1.59 &  14.3 (N=14) \\
        soldier &        0.60 &    0.0 (N=8) \\
project picture &        0.20 &    0.0 (N=3) \\
       soldiers &        0.18 &    0.0 (N=5) \\
      dark room &       -0.47 &    0.0 (N=1) \\
        machine &       -0.95 &   20.0 (N=5) \\
            gun &       -1.69 &  30.0 (N=20) \\
       nice gun &       -1.98 &    0.0 (N=2) \\
    machine gun &       -2.00 &   20.0 (N=5) \\
        weapons &       -2.01 &   42.9 (N=7) \\
\bottomrule
\end{tabular}
  \end{minipage}
  \vspace{3mm}
  \centering 
  \hfill \begin{minipage}{0.45\linewidth} \centering
\begin{tabular}{rll}
\multicolumn{3}{c}{Academic Gown class (ours: ``many people in robes'')} \\
\toprule
            Keyword &  CLIP Score & Subgroup Acc \\
\midrule
             person &        0.51 &  25.0 (N=24) \\
              photo &        0.50 &  27.3 (N=11) \\
           graduate &        0.23 &  13.3 (N=15) \\
          graduates &        0.21 &   25.0 (N=8) \\
         graduation &        0.16 &  13.3 (N=15) \\
               pose &        0.12 &  20.0 (N=10) \\
              poses &        0.01 &    0.0 (N=4) \\
           students &       -0.25 &    0.0 (N=7) \\
graduation ceremony &       -0.57 &  16.7 (N=12) \\
           ceremony &       -1.27 &  23.1 (N=13) \\
\bottomrule
\end{tabular}
  \end{minipage} \hfill
  \begin{minipage}{0.45\linewidth} \centering
\begin{tabular}{rll}
        \multicolumn{3}{c}{Bighorn class (ours: ``rocky hillside'')} \\
\toprule
             Keyword &  CLIP Score & Subgroup Acc \\
\midrule
                goat &        1.39 &   5.9 (N=17) \\
               sheep &        1.35 &    0.0 (N=9) \\
       mountain goat &        0.11 &    0.0 (N=2) \\
          biological &        0.06 &   28.6 (N=7) \\
  biological species &        0.00 &   28.6 (N=7) \\
             species &       -0.03 &   28.6 (N=7) \\
       bighorn sheep &       -0.04 &    0.0 (N=4) \\
bighorn sheep stands &       -0.29 &    0.0 (N=2) \\
              stands &       -1.32 &    0.0 (N=7) \\
                herd &       -2.27 &   14.3 (N=7) \\
\bottomrule
\end{tabular}
  \end{minipage} \hfill
  \vspace{3mm}
  \centering 
  \hfill \begin{minipage}{0.45\linewidth} \centering
\begin{tabular}{rll}
       \multicolumn{3}{c}{Loupe class (ours: ``person holding a magnifying glass'')} \\
\toprule
         Keyword &  CLIP Score & Subgroup Acc \\
\midrule
           black &        0.01 &    0.0 (N=4) \\
          camera &       -0.08 &   28.6 (N=7) \\
            book &       -0.65 &   33.3 (N=3) \\
         compact &       -0.72 &    0.0 (N=2) \\
  compact camera &       -1.12 &    0.0 (N=2) \\
           watch &       -1.50 &    0.0 (N=1) \\
          pocket &       -1.84 &    0.0 (N=1) \\
    pocket watch &       -2.05 &    0.0 (N=1) \\
           glass &       -2.30 &  57.1 (N=14) \\
magnifying glass &       -6.19 &   71.4 (N=7) \\
\bottomrule
\end{tabular}
  \end{minipage} \hfill
  \begin{minipage}{0.45\linewidth} \centering
\begin{tabular}{rll}
  \multicolumn{3}{c}{Weasel class (ours: ``snow weasel'')} \\
\toprule
             Keyword &  CLIP Score & Subgroup Acc \\
\midrule
       bear cub sits &        1.72 &    0.0 (N=1) \\
      black bear cub &        1.62 &   50.0 (N=2) \\
    young black bear &        0.81 &    0.0 (N=1) \\
  biological species &       -0.17 &  85.7 (N=14) \\
      dead squirrels &       -0.19 &    0.0 (N=1) \\
          file photo &       -0.19 &    0.0 (N=2) \\
        undated file &       -0.36 &    0.0 (N=2) \\
  undated file photo &       -0.41 &    0.0 (N=2) \\
               grass &       -0.87 &   85.7 (N=7) \\
squirrels were found &       -0.87 &    0.0 (N=1) \\
\bottomrule
\end{tabular}
  \end{minipage} \hfill
  \vspace{3mm}
  \caption{Comparison to Bias-to-Text~\citep{kim2023explaining}, an automated bias discovery method on ImageNet.
    We show the top 10 keywords identified by Bias-to-Text in descending order of their recommended score.
    We also show the text feedback provided through \ours{} for comparison.
    The keywords identified by Bias-to-Text often include irrelevant words or correspond to very small subpopulations, indicating that current automated methods ultimately require human oversight or intervention to discover the most relevant and biased subpopulations.}
  \label{tab:bias_to_text}
\end{table*}}

\textbf{Datasets.}
We run experiments on three datasets: Waterbirds~\citep{sagawa2019distributionally}, CelebA~\citep{liu2015faceattributes}, and ImageNet~\citep{deng2009imagenet}.
Waterbirds and CelebA have a known spurious correlation between the class label and a spurious attribute; we have access to ground truth spurious attribute labels for these datasets.
We use these datasets to evaluate whether \ours{} can correct model failures due to spurious correlations.
To our knowledge, ImageNet does not have any previously known spurious correlations.

\textbf{Backbone models.}
All experiments use pre-trained CLIP models~\citep{radford2021learning} as the feature extractor.
The \ours{} interface uses the CLIP ViT-L/14 vision and language backbones for calculating image-text similarity.
We use the CLIP ResNet-50 and ViT-L/14 models for Waterbirds and CelebA and only the CLIP ViT-L/14 model for ImageNet.
We use frozen backbone models and only train a final linear layer for classification, following related works for addressing spurious correlations~\citep{kirichenko2022last,zhang2022contrastive}.
We use no data augmentation and normalize all embeddings before computing similarity or training.

\textbf{Methods.}
\cref{table:main_results} and \cref{table:main_results_full} show results for \ours{} and several representative prior methods for addressing spurious correlations.
We experiment with several variants of standard ERM training with a labeled training set: uniform weighting, class-balanced weighting, and ``worst-class'', a DRO-like weighting scheme that adaptively trains on only the class with the highest loss.
We experiment with two variants of training a model with \ours{} annotations: reweighting data so that each of the two slices has equal weight (slice-balanced), and a DRO-like weighting scheme which adaptively trains on only the slice with the highest loss (worst-slice).

\textbf{Annotators.}
We recruit 26 non-expert users through Prolific (https://www.prolific.co/).
These participants had no qualifications beyond being native English speakers and having some programming experience and did not necessarily have any prior knowledge about machine learning.
We provide a brief tutorial on using the interface and ask each participant to annotate the class with the highest error rate for each dataset. 
After completing the user study, we retrained the models for both datasets using each user-provided annotation.
The authors of this paper collected another set of annotations for Waterbirds and CelebA, which we use as a baseline for comparison.
Additionally, annotations for the ImageNet dataset were collected by paper authors.

\section{Additional Details for Bias-to-Text Experiment}
Here, we provide additional details for the comparison between \ours{} and Bias-to-Text~\citep{kim2023explaining}, an automated bias discovery method.
The automated pipeline of Bias-to-Text consists of two steps: (1) extracting keywords from image captions of incorrect examples and (2) ranking these potential keywords based on how well they separate correct and incorrect examples.
More specifically, they look for keywords that maximize CLIP score, which is defined as 
\begin{align}
  s_{\mathrm{CLIP}}(a ; \mathcal{D}):=\operatorname{sim}\left(a, \mathcal{D}_{\text {wrong }}\right)-\operatorname{sim}\left(a, \mathcal{D}_{\text {correct }}\right)
\end{align}
where $\mathcal{D}_{\text {wrong }}$ and $\mathcal{D}_{\text {correct }}$ are the sets of incorrect and correct examples, respectively.
A keyword with a high CLIP score will likely describe something in common between the incorrect examples and thus may correspond to a spurious correlation.
For each keyword, they also report the subgroup accuracy, which is the model's accuracy on the subset of examples containing the keyword.
This method is representative of the state-of-the-art in automated bias discovery and was shown to outperform other recent automated bias discovery methods such as ERM confidence~\citep{liu2021just}, Failure Direction~\citep{jain2022distilling}, and Domino~\citep{eyuboglu2022domino}.

We evaluated the automated pipeline of Bias-to-Text on several classes in the ImageNet validation set in which we identified spurious correlations and found specific pitfalls that make it difficult to use alone in practice.
In Table~\ref{tab:bias_to_text}, we show 10 keywords identified by Bias-to-Text for four of the classes for which we identified spurious correlations.
We note that the top identified keywords, i.e., the ones with the highest CLIP score, often describe something highly related to the class label, such as ``goat'' for the ``bighorn''.
We further note that the method also identifies very small subpopulations, for example ``baby shower'' which only appears in three of the 50 examples in the ``bakery'' class.
Text feedback from \ours{} never the top keyword recommended by Bias-to-Text, and was only in the top 10 for 5 out of 31 classes.

In its current form, automated bias discovery methods such as Bias-to-Text ultimately require oversight to identify the most relevant keywords.
The human-in-the-loop nature of \ours{} can be seen as recognizing this dependency and providing a more direct way for users to inspect and correct model failures.
However, we note that automated discovery methods are still highly useful in the context of \ours{}, as they can prime annotators with a set of candidate keywords or help prioritize the most promising parts of the dataset.
We believe a more integrated combination of automated discovery methods and human-in-the-loop methods such as \ours{} will be a fruitful direction for future work.

\begin{figure*}
    \centering
    \includegraphics[width=0.89\linewidth]{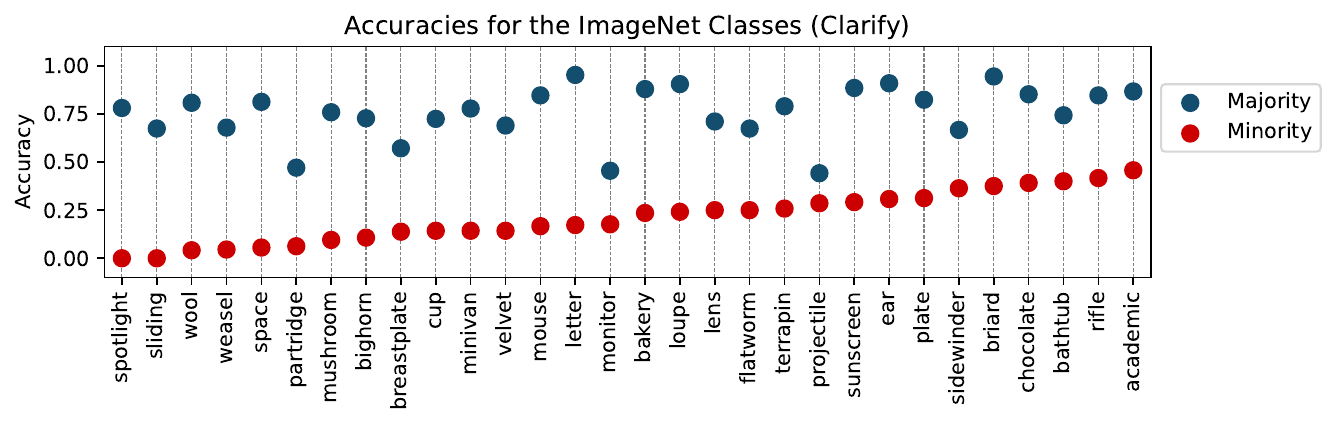}
    \includegraphics[width=0.89\linewidth]{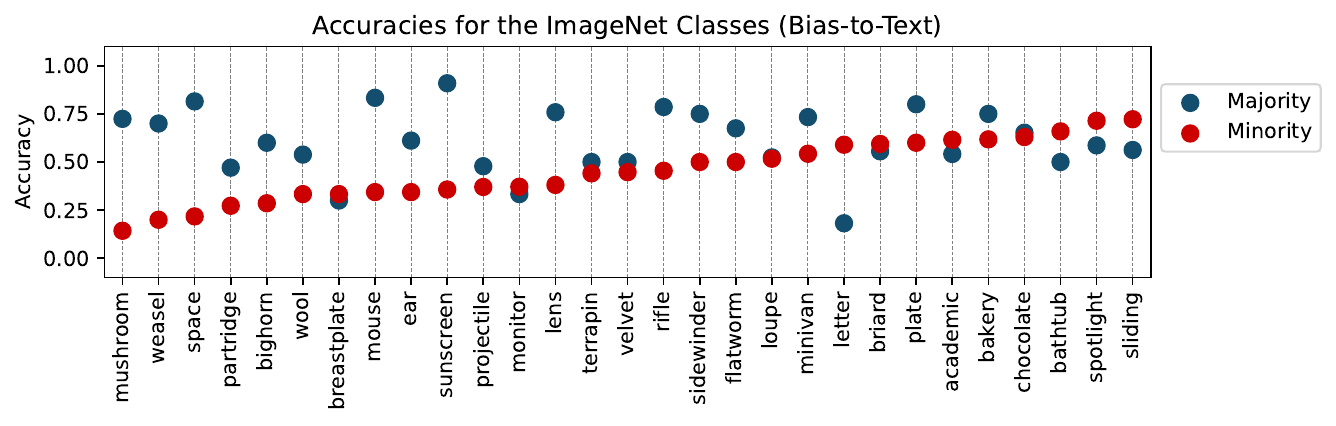}
    \caption{Comparison of annotations discovered by Clarify and Bias-to-Text.We show the accuracy of an ImageNet-trained model on the validation setFor each class with an identified spurious correlation, we show majority split and minority split accuracy.
The annotations for Bias-to-Text show substantially higher minority split accuracy (Clarify: 21.1\%, Bias-to-Text 45.2\%), 
with a smaller gap with the majority split.This indicates that Clarify was substantially more accurate in identifying hard subpopulations.}
\label{fig:b2t_accs_comparison}
\end{figure*}

Finally, we compare the annotations discovered by Clarify and Bias-to-Text.
We take the top keyword identified by Bias-to-Text for each class and compare the model's accuracy on the majority and minority splits in~\cref{fig:b2t_accs_comparison}.
The annotations for Bias-to-Text show substantially higher minority split accuracy (Clarify 21.1\%, Bias-to-Text 45.2\%), with a smaller gap with the majority split.
Furthermore, after re-training with these annotations using our reweighting procedure, we observed a slight decrease in held-out minority split accuracy (45.2\% to 44.3\%).
This is in contrast to re-training with Clarify annotations, which substantially improved minority split accuracy (21.1\% to 28.7\%).
These results indicate that automated bias discovery methods such as Bias-to-Text fail to identify the most relevant or consistent subpopulations, highlighting the need for oversight.

\begin{table*}[!t]
\centering
\begin{tabular}{@{}lllcccccc@{}}
\toprule
& &                      & \multicolumn{3}{c}{Waterbirds} & \multicolumn{3}{c}{CelebA} \\ 
\cmidrule(l){4-6} \cmidrule(l){7-9}
& Assumptions & Method & WG & Avg & Gap & WG & Avg & Gap \\ 
\midrule
\parbox[t]{2mm}{\multirow{18}{*}{\rotatebox[origin=c]{90}{CLIP ResNet-50}}}
& \multirow{2}{*}{Zero-Shot} 
  & Class Prompt & 36.6 & 92.2 & 55.6 & 74.0 & 81.9 & 7.9 \\ 
& & Group Prompt & 55.9 & 87.8 & 31.9 & 70.8 & 82.6 & 11.8 \\ 
\cmidrule{3-9} 
& \multirow{4}{*}{Labels} 
  & ERM & 7.9  & 93.5 & 85.6 & 11.9 & 94.7 & 82.8 \\
& & ERM (ours)                    & 63.4 & 96.0 & 32.6 & 31.1 & 95.4 & 64.3 \\
& & ERM (ours, class-balanced)     & 48.6 & 95.2 & 46.7 & 65.8 & 93.4 & 27.6 \\
& & ERM (ours, worst-class)        & 55.9 & 95.8 & 39.9 & 56.9 & 94.1 & 37.2 \\
\cmidrule{3-9}
& \multirow{3}{*}{Labels, Text Feedback} 
  & \ours{} (avg non-expert)     & 69.8 & 84.1 & 13.3 & 83.7 & 93.2 & 9.5 \\
& & \ours{} (best non-expert)     & \textbf{82.5} & 90.7 & 8.2 & 88.8 & 92.9 & 4.1 \\
& & \ours{} (author)         & 75.7 & 83.8 & \textbf{8.1} & \textbf{89.1} & 92.1 & \textbf{3.0} \\
\cmidrule{3-9} 
& \multirow{4}{*}{Labels, Group Annotation}
  & DFR (downsample)        & 63.9 & 91.8 & 27.9 & 76.9 & 92.5 & 15.6 \\
& & DFR (upsample)         & 51.3 & 92.4 & 41.1 & 89.6 & 91.8 & 2.2 \\
\addlinespace
& & DFR (our implementation) & 78.7 & 90.8 & 12.1 & \textbf{90.6} & 91.9 & \textbf{1.3} \\
& & Group DRO (our implementation) & \textbf{81.3} & 88.1 & \textbf{6.8}  & 89.2 & 91.8 & 2.7 \\
\cmidrule{3-9} 
& \multirow{3}{*}{Labels, Additional Params}
  & ERM Adapter            & 60.8 & 96.0 & 35.2 & 36.1 & 94.2 & 58.1 \\
& & WiSE-FT                & 49.8 & 91.0 & 41.2 & 85.6 & 88.6 & 3.0 \\
& & Contrastive Adapter    & \textbf{83.7} & 89.4 & \textbf{5.7}  & \textbf{90.0} & 90.7 & \textbf{0.7} \\
\midrule
\parbox[t]{2mm}{\multirow{17}{*}{\rotatebox[origin=c]{90}{CLIP ViT-L/14}}}
& \multirow{2}{*}{Zero-Shot} 
  & Class Prompt & 25.7 & 87.3 & 61.6 & 62.1 & 71.9 & 9.8 \\
& & Group Prompt & 27.4 & 85.5 & 58.1 & 72.4 & 81.8 & 9.4 \\
\cmidrule{3-9} 
& \multirow{4}{*}{Labels} 
  & ERM & 65.9 & 97.6 & 31.7 & 28.3 & 94.7 & 66.4 \\
\addlinespace
& & ERM (our implementation)                     & 79.5 & 97.4 & 17.9 & 25.7 & 94.6 & 68.9 \\
& & ERM (our implementation, class-balanced)     & 71.1 & 97.2 & 26.1 & 63.7 & 92.6 & 28.9 \\
& & ERM (our implementation, worst-class)        & 74.3 & 97.1 & 22.8 & 56.9 & 93.3 & 36.4 \\
\cmidrule{3-9}
& \multirow{1}{*}{Labels, Text Feedback} 
  & \ours{} (author)& \textbf{81.8} & 96.8 & \textbf{14.9} & \textbf{88.8} & 90.9 & \textbf{2.1} \\
\cmidrule{3-9} 
& \multirow{4}{*}{Labels, Group Annotation} 
  & DFR (downsample)          & 51.9 & 95.7 & 43.8 & 76.3 & 92.1 & 15.8 \\
& & DFR (upsample)           & 65.9 & 96.1 & 30.2 & 83.7 & 91.2 & 7.5 \\
\addlinespace
& & DFR (our implementation) & 85.9 & 93.5 & 7.6 & \textbf{89.0} & 90.9 & \textbf{1.9} \\
& & Group DRO (our implementation) & \textbf{88.5} & 92.7 & \textbf{4.1} & 88.1 & 91.1 & 2.9 \\
\cmidrule{3-9} 
& \multirow{3}{*}{Labels, Additional Params} 
  & ERM Adapter            & 78.4 & 97.8 & 19.4 & 36.7 & 94.2 & 57.5 \\
& & WiSE-FT                & 65.9 & 97.6 & 31.7 & 80.0 & 87.4 & 7.4 \\
& & Contrastive Adapter    & \textbf{86.9} & 96.2 & \textbf{9.3} & \textbf{84.6} & 90.4 & \textbf{5.8} \\
\bottomrule
\end{tabular}
\caption{
\label{table:main_results_full}
Evaluation of methods for improving group robustness of CLIP models.
Grouped by data and expressivity, with best worst-group (WG) and robustness gaps \textbf{bolded}.
All metrics are averaged over three seeds.
}
\end{table*}

\begin{figure*}[t]
\centering
\includegraphics[width=0.8\linewidth]{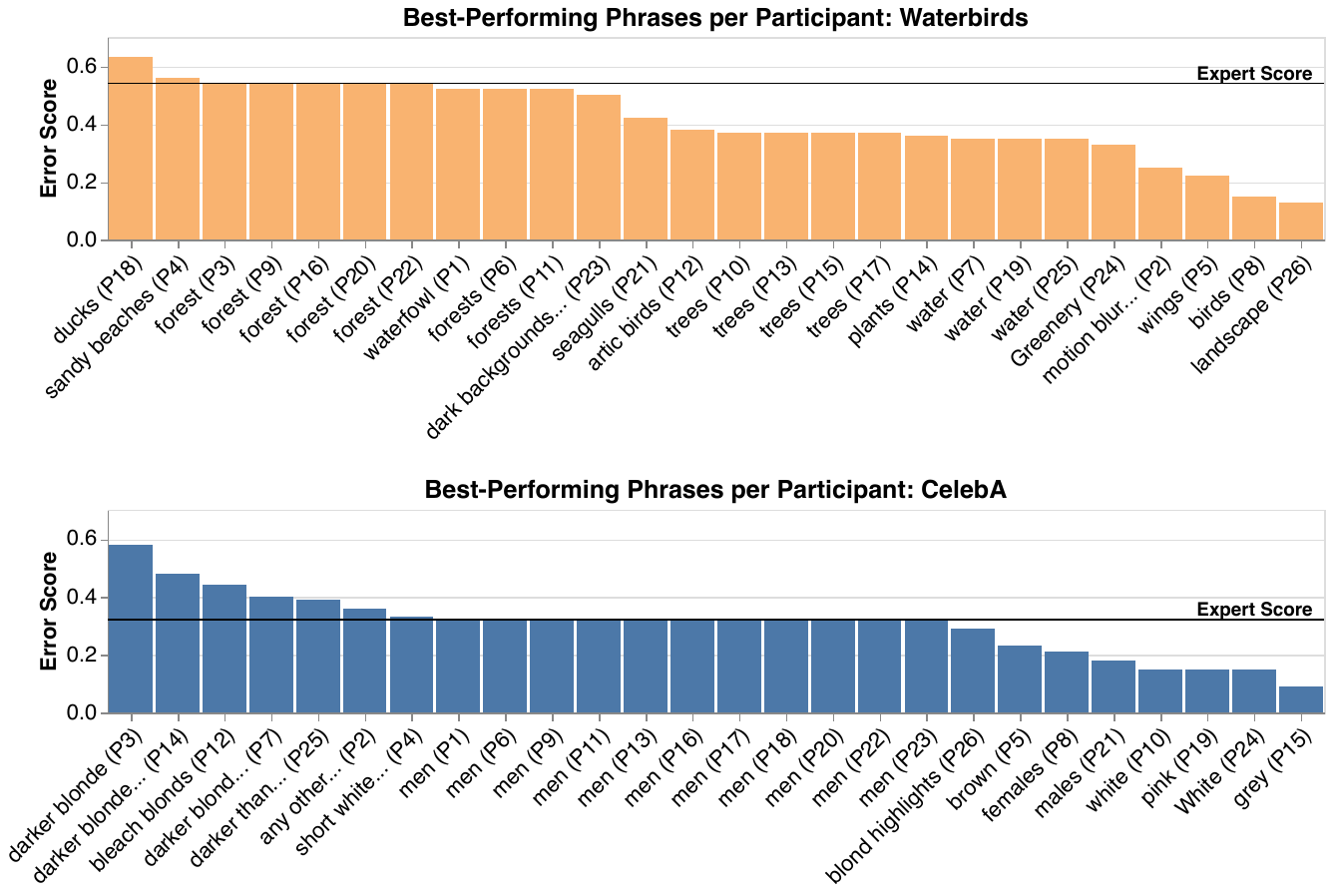}
\label{fig:nonExpert-eval-errorScores}
\caption{Non-experts used \ours{} to identify high-quality descriptions with Error Scores that matched or exceeded the authors' expert annotations.}
\end{figure*}

\begin{figure*}
    \centering
    \includegraphics[width=\linewidth]{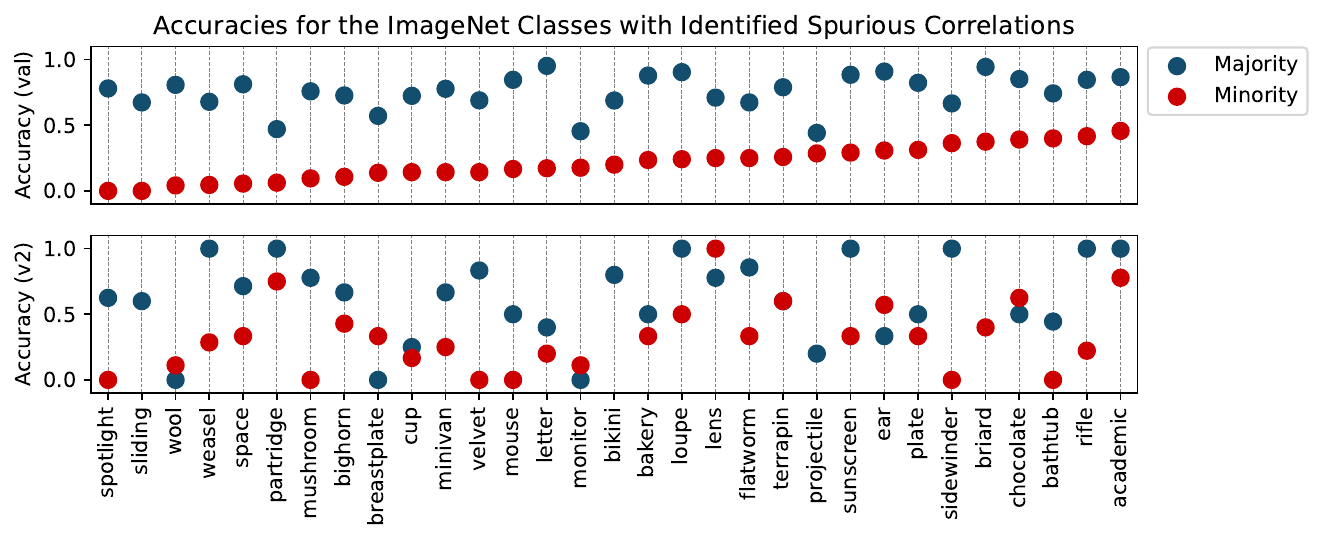}
    \caption{ Accuracy of a model trained on the ImageNet train set, on the ImageNet validation set (top), and on ImageNet-V2 (bottom).
    For each class with an identified spurious correlation, we show majority and minority split accuracy.
    The model achieves lower accuracy on the minority split for all classes in the validation set and all but 6 classes in ImageNet-V2, indicating that the model relies on each identified spurious feature.}
    \label{fig:imagenet-eval}
\end{figure*}

\section{Qualitative Feedback}
We collected qualitative feedback from the non-expert participants in our user study.
We found that the interface was easy to use and provided useful insights into the strengths and weaknesses of the method in~\cref{tab:participant-feedback}.
\begin{table*}[!t]
\centering
\footnotesize
\begin{tabular}{p{0.95\textwidth}}
\toprule
\textbf{Clarity} \\
> Study seemed a little confusing \\
> It was a little confusing at first, but then became clear and challenging. \\
> Compared to the example task, they were a bit more difficult but still understandable. \\
> Yes I understood what was being measured. \\
\midrule
\textbf{Strategies and Thought Processes} \\
> Yes, to find a pattern among the red (incorrect) squares and determine a relevant phrase or description that captures the similarity \\
> Yes. I just tried to figure out what characteristics mainly led to something wrong. \\
> I was honestly having fun trying different prompts. \\
> It was clear in that I had to try to figure out the AI's weakness, but finding that weakness was hard. I tried to do related topics, then used what I saw in the pictures for fodder. \\
> Yes, I would look for what was different about the incorrect images and enter my first guess then work from there. \\
> I was trying to spot what the common things were that the AI was struggling to pick up in the photos it was getting wrong. \\
> The study seemed to be about helping correct the misbehavior of AI. My thought process was mostly linked to trying to find shared features that miscategorized images included. \\
> The only thing I wasn't really sure of was how detailed we could be, how many criteria we could give. I tried to keep it low (one of my higher scorers was just \"men\") but sometimes I had to chain them (\"dark backgrounds and a lot of trees\" or similar) to score well. \\
\midrule
\textbf{Difficulty} \\
> The last one was pretty difficult, but I think I saw all the images correctly. For a second I thought one was a young blonde Trudeau with a flipped-up haircut, but realized that wasn't him \\
> more difficult than I thought to come up with various prompts \\
\midrule
\textbf{Suggestions for Improvement} \\
> It might be nice if our most recent guess was highlighted in the table on the left. Every time I'd submit a guess I found my self trying to remember exactly how I'd phrased it and trying to find it in the table to see how well I did. \\
> Look at the different images, try to find commonalities that might be affecting identification \\
\bottomrule
\end{tabular}
\caption{ \label{tab:participant-feedback}
    Open-ended qualitative feedback from participants, grouped by topic.
}
\end{table*}

\end{document}